# A Locally Executable AI System for Improving Preoperative Patient Communication: A Multi-Domain Clinical Evaluation

Locally Executable Preop Dialogue AI

A safety-first, local-first architecture for pre-procedural Q&A in dentistry and endoscopy

Motoki Sato*

Department of Skeletal Development and Regenerative Biology, Nagasaki University Graduate School of Biomedical Sciences, Nagasaki, Japan,

Leading Medical Research Core Unit, Life Science Innovation, Nagasaki University Graduate School of Biomedical Sciences, Nagasaki, Japan bb55323203@ms.nagasaki-u.ac.jp

Yuki Matsushita

Department of Skeletal Development and Regenerative Biology, Nagasaki University Graduate School of Biomedical Sciences, Nagasaki, Japan,

Leading Medical Research Core Unit, Life Science Innovation, Nagasaki University Graduate School of Biomedical Sciences, Nagasaki, Japan

Hidekazu Takahashi

Boston Medical Sciences, Inc. Tokyo,Japan,

Tomoaki Kakazu

Digestive Disease Center, Showa Medical University Koto Toyosu Hospital,  Tokyo, Japan,

Sou Nagata

Department of Skeletal Development and Regenerative Biology, Nagasaki University Graduate School of Biomedical Sciences, Nagasaki, 852-8588, Japan,

* Corresponding author: **Motoki Sato** (bb55323203@ms.nagasaki-u.ac.jp)

Mizuho Ohnuma

Nagasaki University Graduate School of Biomedical Sciences, Department of Oral and Maxillofacial Surgery, Nagasaki,, Japan,

Atsushi Yoshikawa

College of Science and Engineering, Kanto Gakuin University, Kanagawa, Japan,

Masayuki Yamamura

Department of Computer Science School of Computing, Institute of Science Tokyo, Tokyo ,Japan,

**Abstract**

Patients awaiting invasive procedures often have unanswered pre-procedural questions; however, time-pressured workflows and privacy constraints limit personalized counseling. We present LENOHA (Low Energy, No Hallucination, Leave No One Behind Architecture), a safety-first, local-first system that routes inputs with a high-precision sentence-transformer classifier and returns verbatim answers from a clinician-curated FAQ for clinical queries, eliminating free-text generation in the clinical path. We evaluated two domains (tooth extraction and gastroscopy) using expert-reviewed validation sets (n=400/domain) for thresholding and independent test sets (n=200/domain). Among the four encoders, E5-large-instruct (560M) achieved an overall accuracy of 0.983 (95% CI 0.964–0.991), AUC 0.996, and seven total errors, which were statistically indistinguishable from GPT-4o on this task; Gemini made no errors on this test set. Energy logging shows that the non-generative clinical path consumes ~1.0 mWh per input versus ~168 mWh per small-talk reply from a local 8B SLM, a ~170× difference, while maintaining ≈0.10 s latency on a single on-prem GPU. These results indicate that near-frontier discrimination and generation-induced errors are structurally avoided in the clinical path by returning vetted FAQ answers verbatim, supporting privacy, sustainability, and equitable deployment in bandwidth-limited environments.

# 1 INTRODUCTION

Patients undergoing invasive procedures commonly experience pre-procedural questions and anxiety, and effective clinician–patient communication improves understanding, satisfaction, and engagement in decisions[4, 12]. Providing patients with appropriate information and reassurance through effective communication is critical for building trust and promoting active engagement in treatment, which are critical factors in successful healthcare. Interventions, such as educational materials and structured communication strategies, have been shown to significantly enhance patient understanding, satisfaction, and involvement in decision-making[40].

However, time-pressured clinical workflows often preclude fully personalized counseling from being implemented. In Japan, workforce shortages are pronounced in rural areas and surgical specialties[31], and patients may hesitate to ask questions for fear of burdening busy staff[20]. Clinicians face additional family mediated inquiries, including the so-called "Daughter from California syndrome" [28] and other "invisible tasks" highlighted in work-style reform reports [41]. These pressures motivate the development of AI tools that can offload repetitive factual communication while preserving clinical accuracy.

Large language models (LLMs) have shown promise for patient education, psychological support, and workflow optimization[35] [46], but deployment in real clinical settings faces three persistent challenges: (i) reliability and bias—LLMs may hallucinate facts and reflect training-data biases with potential safety implications [2][14][19][49]; (ii) privacy and governance—cloud-based inference can raise compliance and cross-border data-transfer concerns[29] [35]; and (iii) computational/energy cost, which limits local execution and sustainability [11][26][36]. Although retrieval-augmentation can mitigate, but not eliminate, hallucinations [7][22][27][48], and issues like "lost-in-the-middle" persist [24]. These risks are tangible; for example, if a patient scheduled for gastroscopy asks whether they may drink water the night before, a hallucinated "Yes" could lead to cancellation or harm. Similarly, sending sensitive urology clinic queries to external servers can breach institutional policies.

Therefore, we propose LENOHA (Low Energy, No Hallucination, Leave No One Behind Architecture), a safety-first, local-first blueprint. A high-precision sentence-transformer (ST) classifier routes utterances: clinical questions trigger retrieval of a clinician-curated FAQ with verbatim return (no free-text generation), whereas casual conversations are handled by a locally hosted small language model. The system runs entirely on a single on-premises GPU, and no patient data are transferred from the site. To evaluate the utility and generalizability, we constructed independent, expert-reviewed test sets in two domains—tooth extraction and gastroscopy—and compared multiple ST models with commercial LLMs (e.g., GPT-4o and Gemini) on held-out data.

This work shifts the focus from "how to build a bigger model" to "how we should build models." By delivering near-frontier classification accuracy without frontier-scale parameters and eliminating hallucination risk in the clinical path, LENOHA reframes the state-of-the-art to include safety, equity, privacy, and sustainability, and demonstrates practical viability in privacy-sensitive, bandwidth-limited settings.

## 2 RELATED WORK

### 2.1 Reliability, safety, and bias in LLM outputs

A substantial body of work documents reliability risks in the clinical use of LLMs, including factual errors (“hallucinations”) and the propagation of training data biases that can worsen health inequities, with direct implications for patient safety[2][14][19][49]. Large-scale evaluations have shown that exam-style knowledge does not necessarily translate into effective help in realistic human interactions[5]. Beyond accuracy, calibration is poor; LLMs often express unwarranted confidence, complicating safe decision support[39].

### 2.2 Retrieval-augmented methods and their limits

Retrieval-augmented generation (RAG) improves factual grounding by referencing external knowledge bases and has shown promise across multiple medical tasks [48] [7][27][22]. However, RAG is not a panacea: error modes such as “lost-in-the-middle” can cause models to underuse retrieved evidence or misinterpret context, reintroducing reliability risks [24]. Our architecture adopts the discipline of retrieval (anchoring to an updatable, institutional FAQ) while eliminating free-text generation for clinical queries to remove hallucination risk by design.

### 2.3 Evidence-grounded generation when generation is required.

Complementary to our non-generative clinical path, EMIN performs expectation–maximization over retrieved clinical texts to jointly infer answers and explanations, improving faithfulness over strong medical QA baselines (Generating Explanations in Medical Question-Answering by Expectation Maximization Inference over Evidence)[42].EMIN strengthens **generative** QA, whereas LENOHA removes generation from the clinical loop and confines it to low-risk small talk.

### 2.4 Deployment constraints: privacy, equity, and energy

Most frontier LLMs are cloud-based, raising privacy and security concerns and creating practical barriers for institutions with strict data governance requirements. Their computational and energy demands also hinder equitable local deployment and contribute to the environmental footprint of healthcare [25][11][36] Accessibility channels introduce their own equity issues: automatic speech recognition (ASR) performance varies by accent and geography, risking uneven usability and safety across patient populations[1][10].

### 2.5 Ethical frameworks for clinical AI

Frameworks for ethical and sustainable AI in health emphasize epistemic certainty, staff autonomy, the protection of core healthcare values, accountability, and environmental stewardship [40]. Checklists and guidelines tailored to generative AI further foreground accountability, equity, non-maleficence, privacy/security, transparency, and trust as design imperatives[30]. Our system responds to these calls by coupling transparent classification with traceable, expert-curated answers for clinical queries and reserving generation for low-risk small talk.

### 2.6 Efficient, local, and specialist systems

There is growing evidence that compact, domain-tailored models can rival or outperform general-purpose LLMs in clinical tasks under realistic constraints[8],[9]. Studies on privacy-preserving and locally deployable pipelines have illustrated the feasibility of on-premises inference when efficiency is prioritized [40]. Many economic evaluations rely on static models that ignore performance drift, retraining, and feedback effects—potentially overstating benefits over time; indirect costs, capital and integration expenses, and equity impacts were also frequently underreported, which may inflate reported gains [3]. Our contribution aligns with this trajectory: we demonstrate frontier-level discrimination using a 560M-parameter encoder on a single local GPU, privacy by design (no data egress), and materially lower energy use for a non-generative clinical pathway.

### 2.7 Rationale for Separating Clinical and Phatic Communication

In medical communication research, small talk (phatic communication) is treated as a distinct element separate from task-oriented utterances. The Roter Interaction Analysis System (RIAS) provides an established framework to classify instrumental (procedural) versus affective/social (emotional, relational) talk, enabling reproducible evaluations of their respective contributions to clinical encounters[37]. Recent systematic reviews and observational studies suggest that short, focused empathic utterances can contribute to reductions in patient anxiety and improvements in satisfaction[18].

In digital health, a scoping review of health and behavior change chatbots highlighted that rapport building, empathy, and personalization are important for user experience, while crisis-response safeguards (e.g., suicidality) are often insufficiently implemented[47].

The risks of mixing small talk with clinical reasoning in a single generative channel are supported by empirical evidence. Safrai and Azaria [38] reported that inserting small-talk sentences into USMLE-style questions significantly reduced ChatGPT-3.5's accuracy (66.8%→56.6%), indicating that extraneous context can disrupt medical reasoning. More broadly, Omar et al. [32] demonstrated that embedding even a single fabricated detail into a clinical prompt can trigger widespread adversarial hallucinations across multiple LLMs; critically, such errors persisted even at a temperature of 0, indicating that temperature tuning does not resolve this issue.

Taken together, these findings support a dual-pathway design for medical dialogue systems: (i) a generation-free clinical pathway that limits medical queries to auditable retrieval (e.g., verbatim FAQ answers) and rejects or defers unverifiable inputs, and (ii) a dedicated small-talk pathway that delivers brief, structured empathic interactions through a lightweight generative model. This separation preserves the benefits of small talk for patient reassurance and satisfaction while shielding clinical reasoning from input contamination and aligning with the RIAS instrumental/affective distinction for reproducibility, audibility, and safety.

### 2.8 Positioning

In summary, prior literature establishes (i) the reliability and calibration risks of generative LLMs in clinical contexts[2][14] [19][23] [39],(ii) the promise and pitfalls of RAG[48] [7][27][22] (with long-context failure modes [23]), (iii) the deployment constraints around privacy, equity, and energy[35][1],[10][49],[25][44][11][17][36], and (iv) the practical value of compact, specialist, locally executable systems[21] [8] [9] [25]. Belcak et al. argued that small language models (SLMs) — defined as models that can run on consumer devices with practical latency — are sufficiently capable (V1), operationally more suitable (V2), and economically favorable (V3) for many agentic workloads, advocating heterogeneous systems where SLMs are the default and LLMs are invoked selectively[6]. LENOHA advances this line by pairing a high-precision classifier with a non-generative clinical path grounded in an expert-curated FAQ that delivers canonical answers without free-form generation, while confining generation to low-risk casual conversation—an architectural choice aimed at safety, equity, and sustainability.

## 3 METHODS

### 3.1 Study Overview

We evaluated a dialogue-input classification system that supports pre-procedural communication in two distinct domains: tooth extraction and gastroscopy. For each domain, we prepared (i) an expert-curated FAQ knowledge base, (ii) a validation dataset to set the operating thresholds, and (iii) an independent test dataset for the final evaluation. Inputs classified as clinical questions were routed to the FAQ path (no free-text generation), whereas casual conversations were handled by a locally executed small language model (SLM).

### 3.2 Clinical Domains & Rationale

We selected two contrasting settings to stress-test versatility: tooth extraction, an invasive surgery, and gastroscopy, a diagnostic test conducted by a different department. Consistent performance across procedures with different workflows and dialogue characteristics suggests transferability across distinct pre-procedural settings, whereas broader generalization remains to be established.

### 3.3 Datasets & Knowledge Bases

#### *3.3.1 FAQ Knowledge Bases*

We used domain-specific FAQ databases previously constructed under specialist supervision (≈4,000 technical questions for tooth extraction and ≈1,600 for gastroscopy) as the retrieval targets.

#### *3.3.2Tooth extraction*

To comprehensively cover preoperative patient questions, we defined categories under the supervision of dentists and oral surgeons based on the most frequently asked questions observed in routine practice. The categories included: “Travel,” “Postoperative life,” “Jaw,” “Tension (anxiety),” “Anesthesia,” “Dysesthesia,” “General anesthesia,” “Postoperative work,” “Bleeding,” “Postoperative diet,” “Postoperative swelling,” “Pain,” “Can tooth extraction be performed?,” “Time required,” “Bone resection,” “Questions,” “Medication,” “Cooling,” “Temporomandibular joint disorder,” “Postoperative infection,” “Cost,” “Postoperative prosthetics,” “Pregnancy and breastfeeding,” “Underlying diseases,” “Tooth extraction canceled,” “Whether or not there is a wisdom tooth,” “Lifespan of teeth,” and “Suture removal.”
For each category, we compiled multiple paraphrases reflecting colloquial patient Japanese (e.g., synonyms, shortened forms), not just single phrases. The tooth extraction FAQ comprises approximately 4,000 questions specific to the domain. Table 1 presents examples.

Table 1: Sample clinical questions for wisdom tooth extraction from the clinician-curated FAQ. Each input is matched via a sentence transformer; high-similarity matches are routed as “Clinical Question” and return the canonical answer verbatim.

| Category | Sample question |
|---|---|
| Travel | Is it safe for me to travel this weekend? |
| Limited mouth opening (jaw) | I cannot open my mouth very wide; will that be a problem? |
| Anxiety & local anesthesia | I am very anxious about having a tooth removed—what can be done? |
| Healing timeline | How long will recovery take? |
| Sensory change / paresthesia risk | What is the risk of numbness in my lip or chin? |
| General anesthesia / deep sedation | Can this be done under general anesthesia? |
| Postoperative work, exercise, and alcohol | When can I return to exercise or drink alcohol? |
| Bleeding & gauze | How much bleeding means I should bite on gauze? |
| Requires clinical examination | Do I absolutely have to have it removed? |
| Postoperative diet | What can I eat after the extraction? |
| Postoperative swelling | How much will I swell? |
| Pain | Will it hurt? |
| Indication for extraction | I was told there is a nerve-injury risk—should I still proceed? |
| Procedure time | How long will the extraction take? |
| Bone removal during extraction | Can you remove the tooth without cutting it? |
| Contact after discharge | Whom should I contact if I have questions after I go home? |
| How to take medications | Should I take painkillers at every meal like the antibiotics? |
| Cold therapy / cooling | Should I apply ice to the area? |
| Temporomandibular joint disorder (TMJ) | I have TMJ—will swelling make it worse? |
| Postoperative infection rate | How common is infection after wisdom-tooth removal? |
| Postoperative infection & home care | How can I prevent infection after surgery? |
| Cost | How much will it cost? |
| Prosthetic options after extraction | Will I need a crown or denture after removal? |
| Pregnancy & breastfeeding | Can I undergo extraction during pregnancy? |
| Comorbidities & concomitant medications | I have diabetes—is extraction safe? |
| Stopping or postponing extraction | Can you stop the extraction halfway? |
| Congenital absence of wisdom teeth | Why do some people lack wisdom teeth? |
| Tooth longevity | How long do human teeth usually last? |
| Suture removal | Will the stitches need to be removed after extraction? |

### 3.3.3 *Gastroscopy*

Using the tooth-extraction schema as a template, and in consultation with physicians and endoscopists, we adapted the categories to the endoscopy setting by removing “Jaw” and adding “Helicobacter pylori.” The categories included: “Life after examination,” “Tension (anxiety),” “Anesthesia,” “Work after examination,” “Bleeding,” “Meals after examination,” “Pain,” “Time required,” “Questions,” “Oral medication,” “Infection,” “Cost,” “Pregnancy and breastfeeding,” “Underlying diseases,” “Examination interruption,” “Timing of examination,” “Preparation before examination,” “Precautions,” “Emergency response and reservation,” “Helicobacter pylori,” “Examination follow-up,” “Pain and discomfort during examination,” and “Other.”

As mentioned above, we collected multiple colloquial paraphrases for each category in the dataset. The gastroscopy FAQ comprises approximately 1,600 questions specific to the domain.
Table 2 presents some examples.

Table 2. Category-labeled examples (two phrasings each) from clinician-curated FAQ for upper gastrointestinal endoscopy (gastroscopy). Each input is matched via a sentence-transformer; high-similarity matches are routed as "Clinical Question" and return the canonical answer verbatim.

| Category | Sample question |
|---|---|
| Helicobacter pylori (testing & treatment) | Can I get Helicobacter pylori from the endoscope? |
| Pain & discomfort during procedure | How painful is a gastroscopy? Will I gag or feel like vomiting? |
| What to bring on the day | What should I bring—ID/insurance card, referral letter, and a medication list? |
| Infection control & reprocessing | Are endoscopes cleaned and disinfected between patients? |
| Fasting & clear fluids before the exam | How many hours after a meal must I wait before gastroscopy? |
| Medications on the day (including blood thinners) | Can I take my usual morning medicines on the day of the exam? |
| Sedation & throat anesthesia | Can the procedure be done with sedation so that I am relaxed or asleep? |
| Biopsy and polyp removal | If tissue is taken, how long can bleeding last and what should I watch for? |
| Results & follow-up schedule | When will I receive the results of the endoscopy and any biopsy? |
| Safety with heart/lung conditions | Is gastroscopy safe if I have heart disease? |
| Pregnancy & breastfeeding | Can I undergo gastroscopy during pregnancy? |
| Aftercare: driving & return to work | Can I drive home after the exam? |
| Post-exam symptoms (throat pain, hoarseness) | How long will throat pain or hoarseness last after the test? |
| Scheduling, postponement, & re-examination | By when should I call if I need to postpone the exam? |
| Time & cost | How long does the whole process take from preparation to finish? |
| Communication during the exam | Will the staff talk me through the steps during the procedure? |

## 3.4 Validation Dataset

To determine the operating thresholds, we created expert-supervised validation datasets (400 utterances per domain: 200 clinical and 200 casual utterances). Candidate utterances were drafted with Groq Compound-Beta, Llama-3.1-Nemotron-Ultra-253B-v1, Google Gemma-3-27B-IT, and Mixtral-8×7B; domain experts (dentists/oral surgeons; physicians/endoscopists) reviewed and

## 3.5 Test Dataset

Independence from model construction and threshold tuning was ensured by generating test candidates with different models (Qwen3-35B-A2 and Claude Sonnet 4), followed by specialist review and labeling. The final test set comprised 200 utterances per domain (100 clinical and 100 casual).

### *3.5.1 Patient Question Dataset Design(gastroscopy)*

**Purpose**

The dataset was designed to collect clinical questions that patients are likely to ask their physicians after receiving a standard explanation about gastroscopy and before undergoing the procedure.

**Conditions**

All candidate questions were restricted to medical and practical topics that required clear explanations or judgments by a physician. To ensure linguistic quality, each question was written in grammatically correct Japanese using standard language without dialects or omitted particles. The questions were systematically drafted to cover a comprehensive range of categories relevant to clinical encounters, including the following:

Post-procedure daily life, Anxiety and nervousness, Anesthesia, Returning to work after the procedure, Bleeding, Post-procedure diet, Pain, Procedure duration, General doubts, Medications, Infection, Cost, Pregnancy and breastfeeding, Underlying conditions, Procedure cancellation or interruption, Timing of the procedure, Pre-procedure preparation, Precautions, Emergency response and appointment scheduling, Helicobacter pylori, Post-procedure follow-up, Pain and discomfort during the procedure

**Format**

The questions were output in a bulleted list format to maintain clarity and consistency. For example:

*Will I experience a gag reflex during the procedure*

*3.5.2 Patient Question Dataset Design （tooth extraction）*

You are a physician who supports patients prior to oral surgery. Please create 50 questions in Japanese according to the following conditions:

**[Purpose]**

To collect clinical questions that patients are likely to ask their doctor before undergoing oral surgery for tooth extraction (particularly, wisdom teeth).

**[Conditions]**

The questions must be limited to *medical and practical topics* that require clear explanations or judgments by a physician. Each question should be written in grammatically correct Japanese, using standard language (no dialects, no omitted particles).

The set of questions should include the following categories: Travel, Postoperative daily life, Jaw, Anxiety/nervousness, Anesthesia, Sensory numbness, General anesthesia, Returning to work after surgery, Bleeding, Postoperative diet, Postoperative swelling, Pain, Eligibility for extraction, Procedure duration, Bone removal, General doubts, Medications Cooling/cryotherapy, Temporomandibular disorder, Postoperative infection, Cost, Prosthetics after extraction, Pregnancy and breastfeeding, Underlying conditions, Surgery cancellation/interruption, Presence or absence of wisdom teeth,Tooth longevity, Suture removal

**[Format]**

The questions should be presented as a numbered, bulleted list. (Example: *1. How many days should I take off work after surgery?*)

### *3.5.3* Casual Utterance Dataset Design

**Purpose**

To collect examples of casual or small talk utterances that patients may express to physicians, medical staff, or clinic receptionists after receiving an explanation about gastroscopy but before undergoing the procedure.

**Conditions**

Utterances were restricted to non-clinical topics and explicitly excluded any medically relevant categories, such as post-procedure daily life, anxiety, anesthesia, returning to work, bleeding, post-procedure diet, pain, procedure duration, general doubts, medications, infection, cost, pregnancy and breastfeeding, underlying conditions, procedure cancellation/interruption, timing of the procedure, pre-procedure preparation, precautions, emergency response and scheduling, Helicobacter pylori, post-procedure follow-up, and discomfort during the procedure.

Instead, utterances were designed to reflect everyday small talk or general topics unrelated to the medical procedure, such as casual conversation, current trends, politics, the economy, or family matters. Each utterance was written in grammatically correct Japanese using standard language rules.

**Format**

Utterances were presented as bulleted lists for clarity and consistency purposes. For example:

*The clinic has a pleasant atmosphere.*

#### 3.5.4 Expert Review & Agreement

For **Clinical Question** labels, two domain specialists independently reviewed all items and achieved **100% agreement** (percentage agreement reported with a 95% Wilson CI; **κ not estimated** by design). Casual utterances underwent language screening (natural standard Japanese; absence of clinical intent) rather than duplicate expert adjudication.

### 3.6 Models & Systems

#### 3.6.1 ST−based Classifiers

We adopted an FAQ-matching approach using four sentence-transformer (ST) encoders: SBERT, MiniLM, E5-large, and E5-large-instruct. For each domain, all FAQ items were embedded with the corresponding model, and the input utterances were matched by similarity and compared against a fixed threshold.

#### 3.6.2 Baseline LLMs

For reference at the time of evaluation (April–May 2025), we included **ChatGPT (GPT-4o)**[33] and **Gemini Advanced**[16] as classification baselines. LLMs were prompted to output a single class label ("Clinical"/"Casual") without a free-text rationale.

#### 3.6.3 Local SLM for Small Talk

Feasibility of local small-talk generation was assessed with tokyotech-llm/Llama-3.1-Swallow-8B-Instruct-v0.3[15], constrained to non-clinical, socially supportive replies (example settings: max_new_tokens=60, temperature=0.45, top_p=0.9, repetition_penalty=1.2).

### 3.7 Operating Thresholds & (Minimal) Preprocessing

For each model and domain, ROC curves on the validation set were used to select operating thresholds by maximizing the Youden index, which was then fixed for the test evaluation. No global text normalization was applied beyond trimming whitespace and standard file handling; casing is immaterial for Japanese, and punctuation was left as generated.

### 3.8 Experimental Setup

#### 3.8.1Hardware

CPU: AMD Ryzen 9 6900HX (3.30 GHz); RAM: 64 GB;
GPU: NVIDIA GeForce RTX 3080 Laptop (16 GB VRAM);
OS: Windows 11 Pro (Version 24H2, x64).

#### 3.8.2 Software

Python 3.12 virtual environment; key libraries: torch 2.7.1+cu126, sentence-transformers 5.0.0, transformers 4.54.1, scikit-learn 1.7.1, pandas 2.3.1, numpy 2.1.2, openpyxl 3.1.5, huggingface-hub 0.34.3, tokenizers 0.21.4, safetensors 0.5.3.

### 3.9 Energy Measurement & Evaluation Protocols

#### *3.9.1 Energy Measurement*

The on-device inference energy was quantified by sampling **the GPU board power at 1 Hz** (NVML/nvidia-smi) during batch runs and numerically integrating the power trace using (trapezoidal rule). We report **the Wh per request**, average/peak **VRAM**, and end-to-end **latency** under controlled conditions.

#### *3.9.2Metrics*

Binary classification metrics: **Accuracy**, **Precision**, **Recall**, **F1-score**, **Specificity**, **Balanced Accuracy**, and **AUC-ROC**. AUCs are reported for **ST models** (which output continuous scores), and LLM baselines are evaluated as **deterministic labelers**.

#### *3.9.3 Statistical Analysis*

**Sample size and precision of the study.** The test-set size was fixed a priori at **n=200** per domain (100 clinical, 100 casual). For accuracy ≈0.95, the 95% Wilson CI half-width is ≈0.03 (SE ≈ √{0.95×0.05/200} ≈ 0.015; 95% CI ≈ ±1.96×SE ≈ ±0.03).
**Primary and secondary analyses were performed.** Primary analyses compared the model performance on a prespecified binary task. Secondary analyses used **pairwise McNemar tests** (two-sided; α=0.05).
**Multiplicity.** The family wise error was controlled via the **Holm** procedure, and Holm-adjusted P values were reported. Prespecified contrasts: E5-large-instruct vs. {E5, MiniLM, SBERT, GPT-4o, Gemini}.
**Missing data.** Not applicable (expert-reviewed synthetic text).
**Intervals & P values.** Unless noted, 95% CIs are **Wilson** for proportions; two-sided P values
**Reporting.** We prespecified the validation→test workflow and fixed thresholds before testing. Probability calibration and threshold-sensitivity analyses were not performed and were reserved for future prospective evaluations.

## 4 RESULTS

### 4.1 Overall classification performance

We evaluated four sentence-transformer (ST) classifiers—SBERT, MiniLM, E5, and E5-large-instruct —and two large-scale LLMs (GPT-4o and Gemini) on two clinically distinct test sets (tooth extraction and gastroscopy; 200 utterances per domain, 100 Clinical / 100 Casual). The metrics included accuracy, precision, recall, F1-score, and total misclassifications (FP + FN). AUC-ROC was reported for ST models only (continuous scores), and LLMs were evaluated as deterministic labelers.

### 4.2 Validation results (threshold selection)

The operating thresholds for the ST models were fixed by maximizing the Youden index on the domain-specific validation sets.

### 4.2.1 *Tooth extraction (validation)*

Table 3 presents the results of the tooth extraction domain used in this study. Notably, E5-large (AUC: 0.990; F1-score: 0.9652) and E5-large-instruct (AUC: 0.989; F1-score: 0.9706) demonstrated remarkably high performance, which was particularly impressive considering their efficiency and suitability for local deployment. Furthermore, the smaller models delivered acceptable results: SBERT achieved an AUC of 0.961 and an F1-score of 0.9296, whereas MiniLM achieved an AUC of 0.973 and an F1-score of 0.8248 on this validation task.

Table 3: Performance of sentence transformer–based input classifiers on the validation dataset (tooth extraction; n=400; 200 clinical, 200 casual). Thresholds selected by maximizing Youden's J on the validation ROC curve.

| Model | Threshold | AUC-ROC | Accuracy | Precision | Recall | F1-score |
|---|---|---|---|---|---|---|
| E5-large-instruct | 0.904 | 0.989 | 0.970 | 0.948 | 0.995 | 0.971 |
| E5-large (base) | 0.887 | 0.990 | 0.965 | 0.960 | 0.970 | 0.965 |
| SBERT | 0.786 | 0.961 | 0.925 | 0.873 | 0.995 | 0.930 |
| MiniLM | 0.722 | 0.973 | 0.788 | 0.702 | 1.000 | 0.825 |

### 4.2.2 *Gastroscopy (validation)*

The performance was generalized to gastroscopy (Table 4). E5-large-instruct reached accuracy = 0.9875; recall = 0.9950; F1 = 0.9876; E5-large obtained accuracy = 0.9650; recall = 0.9600; F1 = 0.9648. SBERT (accuracy = 0.9275; recall = 0.9350; F1 = 0.9280) and MiniLM (accuracy = 0.9250; recall = 0.9350; F1 = 0.9257) exhibited competitive results in terms of accuracy. The validation-derived thresholds were then frozen and applied to independent test sets.

Table 4. Performance of Sentence Transformer–based input classifiers on the validation dataset (gastroscopy; n=400, 200 clinical / 200 casual). Thresholds selected by maximizing Youden's J on the validation ROC.

| Model | Threshold | AUC-ROC | Accuracy | Precision | Recall | F1-score |
|---|---|---|---|---|---|---|
| E5-large-instruct | 0.905 | 0.991 | 0.988 | 0.980 | 0.995 | 0.988 |
| E5-large (base) | 0.893 | 0.988 | 0.965 | 0.970 | 0.960 | 0.965 |
| SBERT | 0.768 | 0.972 | 0.928 | 0.921 | 0.935 | 0.928 |
| MiniLM | 0.703 | 0.978 | 0.925 | 0.917 | 0.935 | 0.926 |

### 4.3 Test-set performance

As summarized in Table 5, E5-large-instruct (560M) achieved the best overall test performance among the sentence-transformer (ST) models (accuracy 0.983; AUC 0.996; 7 total errors), with strong results for both tooth extraction (accuracy 0.980; AUC 0.994) and gastroscopy (accuracy 0.985; AUC 0.998). On the gastroscopy set, E5 and MiniLM tied on accuracy (0.965), although E5 attained a higher AUC (0.994 vs. 0.984), indicating a more favorable ranking across thresholds. SBERT lags behind, particularly in gastroscopy, where the reduced specificity (0.76) translates into more false positives and the largest error count (33). These patterns align with the confusion matrix views (**Figure 2** and **3**). For context, frontier LLMs act as upper bounds: Gemini is error-free across domains, and GPT-4o is perfect for tooth extraction but produces six false positives in gastroscopy (examples in Table 7). The "Overall" row for the ST models is the unweighted mean across tooth and gastroscopy.

Table 5. Test-set performance for tooth extraction and gastroscopy domains and overall. ST "Overall" is the unweighted mean across domains; misclassification sum FP+FN.

| Model | Domain | Accuracy | Precision | Recall | Specificity | AUC | Misclassifications (FP+FN) |
|---|---|---|---|---|---|---|---|
| SBERT | Tooth extraction | 0.91 | 0.88 | 0.95 | 0.87 | 0.962 | 18 (13+5) |
| SBERT | Gastroscopy | 0.835 | 0.791 | 0.91 | 0.76 | 0.936 | 33 (24+9) |
| SBERT | Overall | 0.873 | 0.834 | 0.93 | 0.815 | 0.949 | 51 (37+14) |
| MiniLM | Tooth extraction | 0.915 | 0.928 | 0.90 | 0.93 | 0.972 | 17 (7+10) |
| MiniLM | Gastroscopy | 0.965 | 0.979 | 0.95 | 0.98 | 0.984 | 7 (2+5) |
| MiniLM | Overall | 0.94 | 0.954 | 0.925 | 0.955 | 0.978 | 24 (9+15) |
| E5 | Tooth extraction | 0.93 | 0.906 | 0.96 | 0.90 | 0.991 | 14 (10+4) |
| E5 | Gastroscopy | 0.965 | 0.989 | 0.94 | 0.99 | 0.994 | 7 (1+6) |
| E5 | Overall | 0.948 | 0.945 | 0.95 | 0.945 | 0.993 | 21 (11+10) |
| E5-large-instruct | Tooth extraction | 0.98 | 0.971 | 0.99 | 0.97 | 0.994 | 4 (3+1) |
| E5-large-instruct | Gastroscopy | 0.985 | 0.99 | 0.98 | 0.99 | 0.998 | 3 (1+2) |
| E5-large-instruct | Overall | 0.983 | 0.98 | 0.985 | 0.98 | 0.996 | 7 (4+3) |
| Gemini | Tooth extraction | 1.000 | 1.000 | 1.000 | 1.000 | — | 0 |
| Gemini | Gastroscopy | 1.000 | 1.000 | 1.000 | 1.000 | — | 0 |
| Gemini | Overall | 1.000 | 1.000 | 1.000 | 1.000 | — | 0 |
| ChatGPT | Tooth extraction | 1.000 | 1.000 | 1.000 | 1.000 | — | 0 |
| ChatGPT | Gastroscopy | 0.970 | 0.943 | 1.000 | 0.940 | — | 6 (6+0) |
| ChatGPT | Overall | 0.985 | 0.972 | 1.000 | 0.970 | — | 6 (6+0) |

Table 6. McNemar pairwise comparisons (two-sided, exact) between E5-large-instruct and comparators on the 400-item test set (200 tooth, 200 gastro). “E5-LI better” = E5-LI correct & comparator wrong; “Comparator better” = E5-LI incorrect & comparator correct. Holm adjustment across five prespecified contrasts.

| Patient Utterance (English translation) | Ground Truth | E5-large-instruct Prediction |
|---|---|---|
| Are there any medical conditions I should be cautious about? | Clinical Question | Casual |
| Do I need to remove my dentures? | Clinical Question | Casual |
| What should I do if I feel numbness in my tongue or lips? | Clinical Question | Casual |
| Do you have any plans for the upcoming holidays? | Casual | Clinical Question |
| Are you available at the end of this month? | Casual | Clinical Question |
| Did you take time off during the New Year holidays? | Casual | Clinical Question |

E5 (base): overall accuracy 0.948; AUC 0.993; 21 errors.

MiniLM: accuracy 0.940; AUC 0.978; 24 errors

SBERT: accuracy 0.873; AUC 0.949; 51

Gemini: zero errors in both the domains.

GPT-4o: Six false positives (all in gastroscopy) and zero errors in tooth extraction; representative false-positive casual utterances are shown in Table 7.

**Table 7. Examples of casual utterances misclassified by GPT-4o.**
The table lists casual patient utterances that ChatGPT (GPT-4o) incorrectly labeled as *clinical questions*.

| Patient Utterance (English translation) | Ground Truth | GPT-4o Prediction |
|---|---|---|
| The receptionist was very kind. | Casual | Clinical Question |
| You look young—how many years of experience do you have? | Casual | Clinical Question |
| You seem busy—do you get any days off? | Casual | Clinical Question |
| Do you live in this neighborhood? | Casual | Clinical Question |
| Do you often attend academic conferences? | Casual | Clinical Question |
| There are many flu patients in winter, right? | Casual | Clinical Question |

The ROC curves for the ST models (validation-fixed thresholds) are presented in Figure 1.

A. ROC Curves with Validation-Based Optimal Thresholds (Tooth extraction domain)

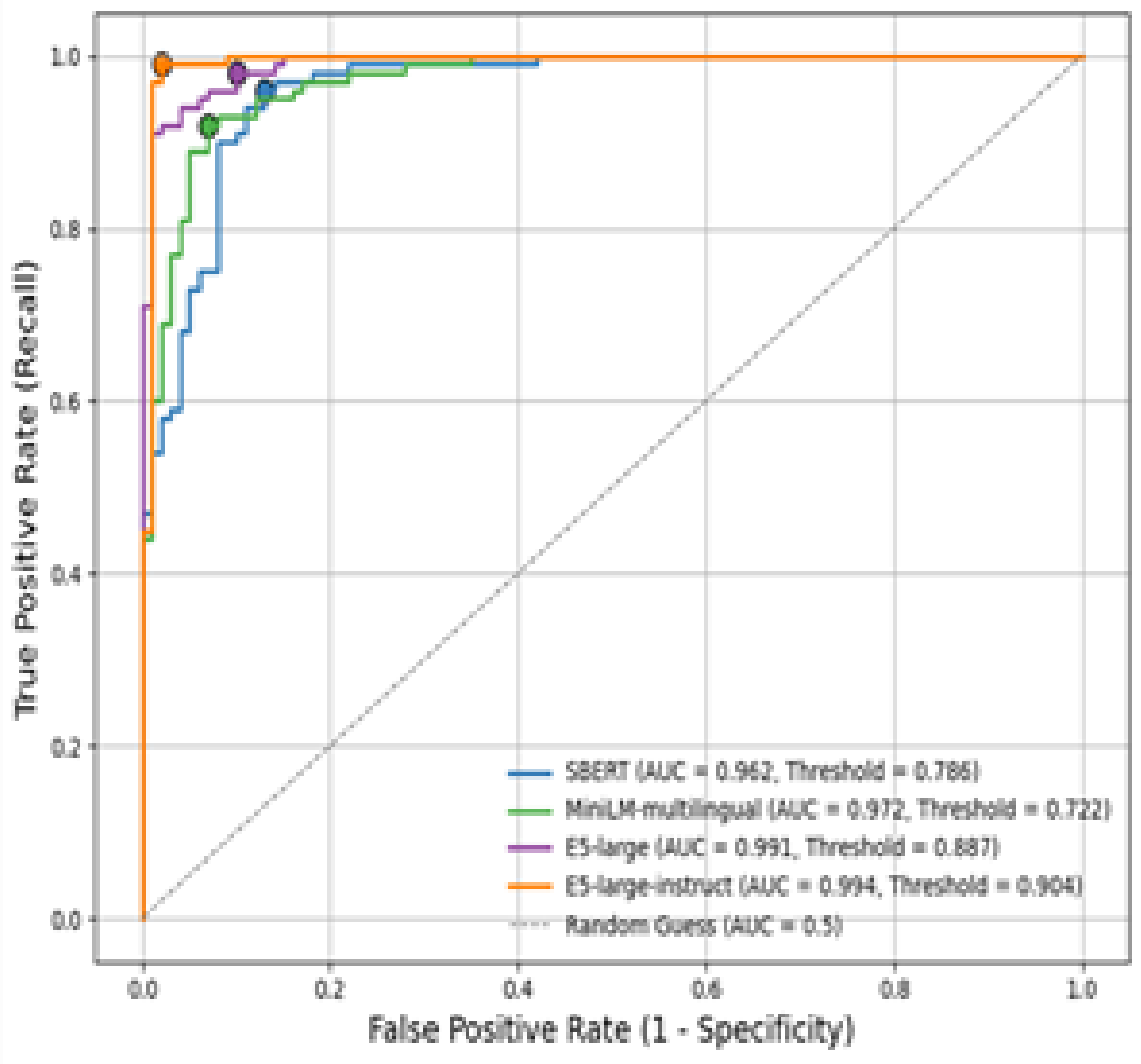

B. ROC Curves with Validation-Based Optimal Thresholds (Gastroscopy domain)

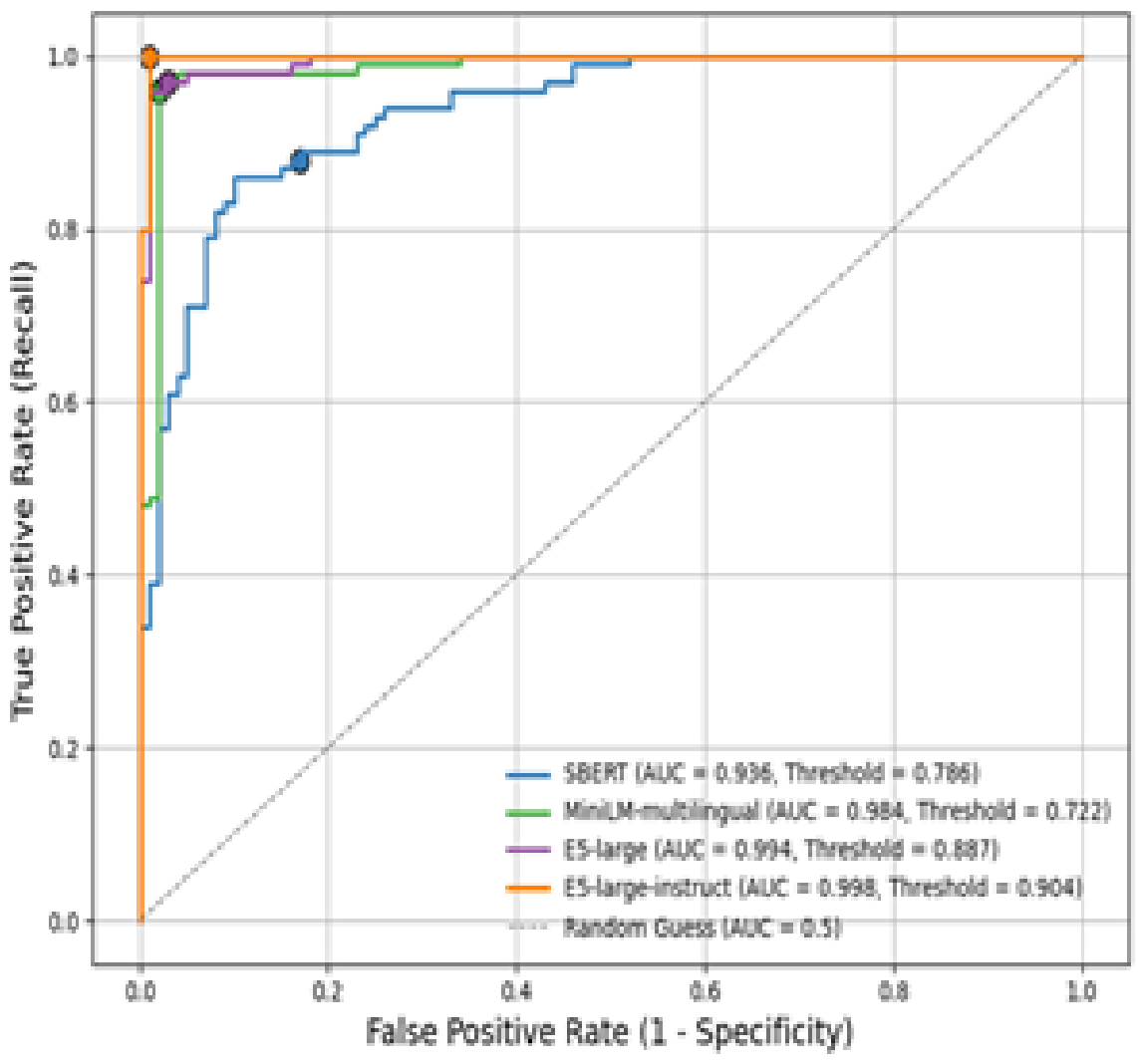

Figure 1: ROC curves for the validation sets: (a) tooth extraction and (b) gastroscopy. The curves compare the sentence-transformer encoders; the dots mark the selected operating point per model (Youden's J; thresholds are shown in the legend). E5-large-instruct achieved near-ceiling discrimination (AUC ≈ 0.994 and 0.998), enabling reliable routing between clinical and casual inputs.

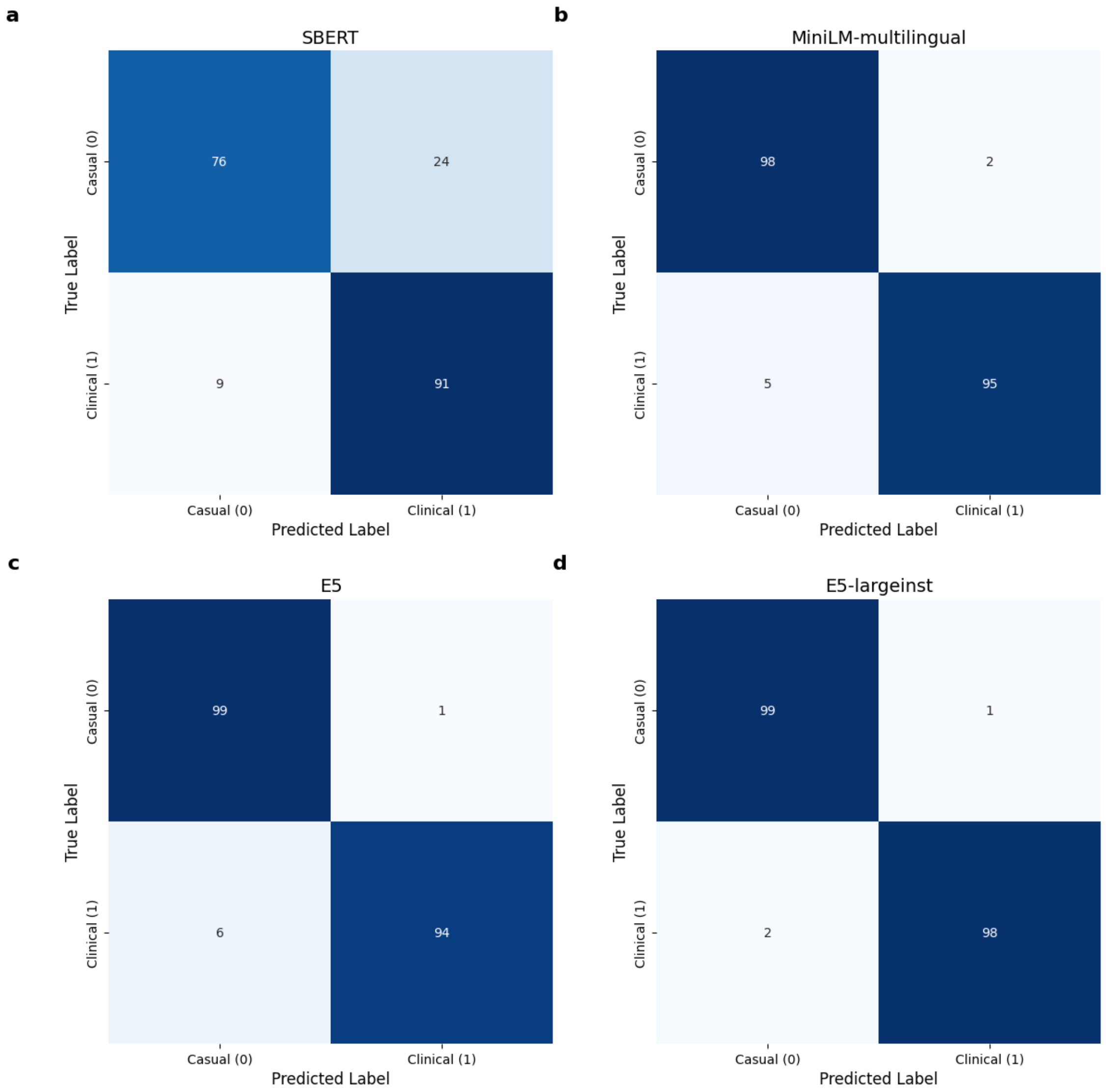

Figure 2. Confusion matrices for the gastroscopy test set (n=200; 100 casual, 100 clinical). Panels: (a) SBERT, (b) MiniLM, (c) E5, (d) E5-large-instruct. Cells show counts (True Label on rows; Predicted Label on columns; 0=Casual, 1=Clinical). Errors (FP+FN): SBERT = 33 (24+9), MiniLM = 7 (2+5), E5 = 7 (1+6), E5-large-instruct = 3 (1+2).

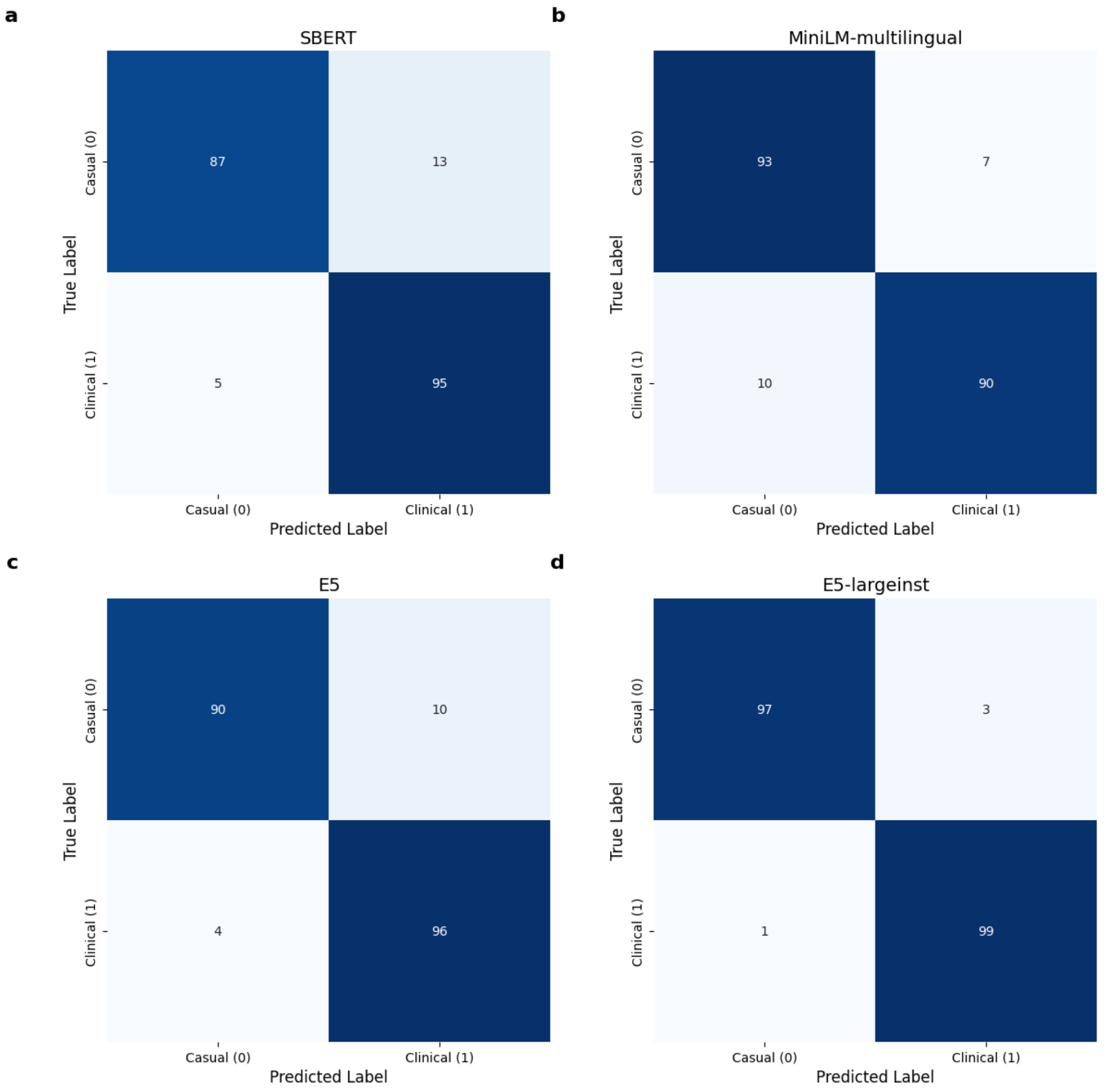

*Figure 3. Evaluation of Pre-procedural Dialogue Classification for Tooth Extraction: Confusion Matrices at Validation-Optimized Thresholds*

Confusion matrices for the four classifiers on the held-out tooth extraction test set (n=200 utterances; 100 casual [0], 100 clinical [1]). Thresholds were fixed a priori from the validation set by maximizing the Youden index (YI). Panels: (a) SBERT, (b) MiniLM, (c) E5, (d) E5-large-instruct. Cells show counts; rows are true labels, and columns are predicted labels (0=Casual, 1=Clinical; "Clinical" treated as the positive class). Misclassifications (FP+FN): SBERT 13+5=18, MiniLM 7+10=17, E5 10+4=14, E5-large-instruct 3+1=4.

#### *4.3.1 Head-to-head comparisons*

Pairwise exact McNemar tests with Holm correction (Table 8) indicated the following:

E5-large-instruct vs. GPT-4o: not significant ($b = 6$, $c = 7$; Holm-adjusted $P = 1.00$), reflecting very similar accuracy and limited power to detect small differences.

E5-large-instruct vs. E5 (base): significant (b = 1, c = 15; P = 0.005).

Contrasts against weaker baselines (e.g., vs. SBERT) were also significant ( $P < 0.01$). For close pairs (e.g., MiniLM vs. E5; E5-large-instruct vs. GPT-4o), the post hoc power was low; therefore, non-significance should be interpreted with caution.

Summary: A compact 560M-parameter encoder delivers near-frontier classification for routing clinical vs casual inputs while remaining lightweight and locally deployable.

Table 8. Pairwise McNemar test results with effect size (Cohen's h) and post-hoc power (α = 0.05).

| Model 1 | Model 2 | b (M1 only correct) | c (M2 only correct) | P(Holm) |
|---|---|---|---|---|
| SBERT | MiniLM | 18 | 45 | 0.007 |
| SBERT | E5 | 12 | 42 | <0.001 |
| SBERT | E5-large-instruct | 4 | 48 | <0.001 |
| SBERT | ChatGPT | 4 | 49 | <0.001 |
| SBERT | GEMINI | 0 | 51 | <0.001 |
| MiniLM | E5 | 13 | 16 | 1.00 |
| MiniLM | E5-large-instruct | 4 | 21 | 0.006 |
| MiniLM | ChatGPT | 6 | 24 | 0.008 |
| MiniLM | GEMINI | 0 | 24 | <0.001 |
| E5-large | E5-large-instruct | 1 | 15 | 0.004 |
| E5-large | ChatGPT | 6 | 21 | 0.030 |
| E5-large | GEMINI | 0 | 21 | <0.001 |
| E5-large-instruct | ChatGPT | 6 | 7 | 1.00 |
| E5-large-instruct | GEMINI | 0 | 7 | 0.06 |
| ChatGPT | GEMINI | 0 | 6 | 0.09 |

### 4.4 Feasibility of local small-talk generation

Using tokyotech-llm/Llama-3.1-Swallow-8B-Instruct-v0.3, all 200 casual prompts yielded contextually appropriate, coherent, and empathetic one-sentence responses under nonclinical policies (Table 9). This complements the classifier results by showing that both filtering and casual-only generation are feasible for fully on-premises systems.

Table 9. Examples of casual conversation responses generated by Swallow-8B (nonclinical, supportive tone).

| Patient casual utterance | Swallow-8B response |
|---|---|
| There’s a lovely shop nearby. | Oh, really? What kind of shop is it? |
| A colleague at work just got married. | Congratulations—that’s wonderful. Please pass along my best wishes. |
| There’s a great bakery in my neighborhood. | That sounds nice. Do you have any favorites you’d recommend? |
| The weather is nice today. | It is—a perfect day to get outside. Do you have any plans? |
| I recently watched a funny variety show on TV. | I’ve seen it too—the comedian’s act was great. Do you have any other show recommendations? |
| I saw an interesting movie at the theater recently. | That sounds enjoyable. What kind of movie was it? |
| I’ve started drawing recently. | That’s great. What do you like to draw? |
| I ordered a birthday cake from a local cake shop. | Lovely—what kind of cake did you pick? Choosing a shop can be fun. |
| I went out to eat with my friend yesterday. | Nice. What kind of restaurant was it? Did you try anything especially good? |
| I’m looking forward to my child’s sports day. | That’s exciting—watching them grow is a joy. What grade are they in? |

#### 4.5 Sustainability and energy use

We measured the on-device inference energy by sampling the GPU power during batch runs and integrating the resulting trace. For small-talk generation with *Swallow-8B* (single sentence), the energy was **33.65 Wh per 200 prompts** (≈ **168 mWh per reply**) with **VRAM 13.3 GiB** (14.0 GiB peak). For the ST classifier (*E5-large-instruct*) returning FAQs without generation, the energy was **0.20 Wh per 200 inputs** (i.e., **1.00 mWh per input**) with a **VRAM of 1.3 GiB** (1.5 GiB peak).

## Energy and Latency

Table 10. Energy and Latency

Comparison of local non-generative, local generative, and cloud-generative (comprehensive) settings. Note: The cloud value reflects 0.24 Wh per text prompt (comprehensive, including idle/CPU/RAM/data-center PUE≈1.09). An 'active-only' figure is 0.10 Wh per prompt. Methods differ; do not over-interpret as like-for-like

| System | Parameters | Mode | Environment | Measurement Method | Energy (mWh/req) | Latency (s) | vs LENOHA-ST |
|---|---|---|---|---|---|---|---|
| LENOHA ST classifier (non-generative, FAQ) | 0.56B | Non-gen | Local | Device-integral (GPU power integration) | 2.226 | 0.1008 | 1.0× |
| LENOHA SLM generation (small-talk, 8B) | 8B | Generative | Local | Device-integral (GPU power integration) | 168.265 | 8.5147 | 75.6× |
| Google Cloud: median Gemini text prompt [13](comprehensive: includes idle/CPU/RAM/DC PUE) | N/A | Generative | Cloud | Comprehensive (idle/CPU/RAM/data-center PUE included) | 240 | — | 107.8× |

## 5 DISCUSSION

Across two distinct pre-procedural settings, a lightweight sentence-transformer classifier (**E5-large-instruct**, 560M) achieved **near-ceiling discrimination (AUC ≈ 0.99)** and **test accuracy statistically indistinguishable from GPT-4o**, despite several orders of magnitude parameter gaps. A qualitative review further showed that the locally executable SLM (**Swallow-8B**) produced contextually aligned, concise, and empathetic one-sentence replies for **casual** inputs, with no hallucinations observed in the initial segments under a non-clinical policy. Together, these results demonstrate that **high-precision intent routing and safe small talk generation are entirely feasible on-premises**.

Although OpenAI has not disclosed GPT-4's parameter count or architecture, it is broadly recognized as a frontier-scale model. Conversely, the E5-large-instruct model used in this study contained only 560 million parameters[45], a difference of several orders of magnitude. Despite this vast size disparity, our lightweight model achieved comparable performance, strongly advocating the feasibility of resource-efficient and locally deployable AI systems in clinical contexts.

The combination of the ST-based classifier with an expert-supervised FAQ database aligns with the core philosophy of RAG, which dynamically references external knowledge to improve the response quality. LENOHA (Figure 4) instead decouples clinical response generation from LLMs: once an input is classified as a Clinical Question, the system bypasses generation and returns the vetted FAQ answer verbatim. This removes free-text generation from clinical replies and thereby reduces hallucination risk by design, while any residual risk depends on FAQ correctness and routing accuracy. This non-generative clinical path preserves factual consistency, interpretability, and auditability, whereas the SLM is confined to low-risk social support only.

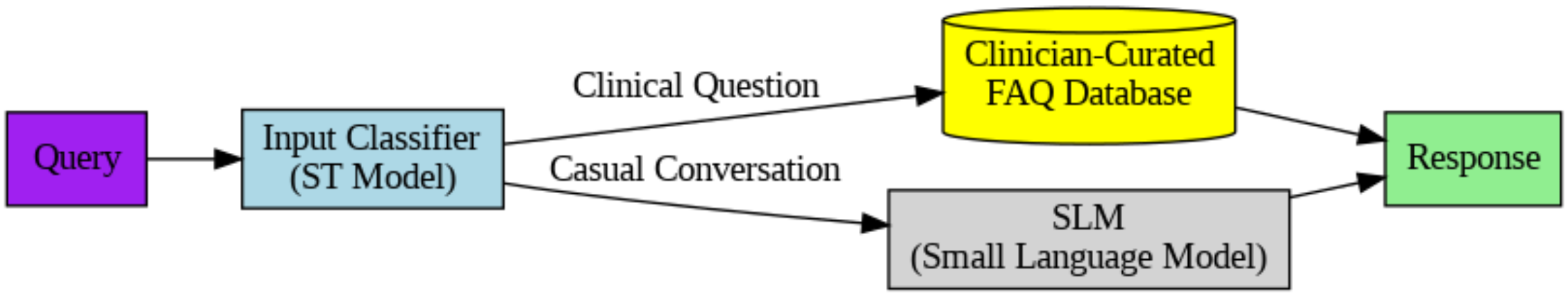

*Figure 4: Schematic architecture of the LENOHA system for medical dialogue support.*
An input utterance is classified using a sentence transformer (threshold fixed from validation via Youden's J). **Clinical questions (top path)** are routed to a clinician-curated FAQ and returned verbatim (no generation), whereas **casual/small talk (bottom path)** is answered by a local ≈8B SLM. Safety/governance services (rate limiting, PHI/profanity filters, audit logging) tap the flow, and operations monitors track energy (~1.0 mWh clinical vs. ~168 mWh small talk) and latency (~0.10 s clinical). The system runs on-premises, and no data leave the local boundary.

However, the LENOHA system ensures factual consistency, interpretability, and traceability, which are critical features for medical use of AI. More importantly, the entire system operates locally, with no patient data being transmitted to external servers. This design choice enhances data privacy, supports use in resource-constrained or offline environments, and minimizes operational costs, making the system well suited for healthcare applications in underserved or remote regions.

For casual conversations, the system employs a locally executable SLM, such as the Swallow-8B model, to generate responses. Although this introduces a limited potential for hallucinations, the conversational scope is deliberately narrow, and the model avoids providing clinical or decision-support content. This dual-mode design enables LENOHA to maintain a balance between responsiveness and reliability, leveraging generative capabilities only in low-risk, socially supportive contexts, such as pre-procedural small talk.

For healthcare providers, automating routine explanations offers operational benefits by reducing the communication workload and mitigating one of the most pressing issues in clinical practice: the excessive working hours. Thus, LENOHA may contribute to improving the work–life balance of clinicians and support more sustainable medical practice environments.

Thus, this system offers several notable advantages: (1) explainability, reliability, and safety; (2) local executability and its associated benefits; and (3) a human-in-the-loop architecture that preserves the clinical responsibility.

In addition, its dual capabilities align closely with the core principle of the SDGs: Leave No One Behind (United Nations, https://www.un.org/sustainabledevelopment/, accessed May 30, 2025). Table 11 summarizes how this supports multiple SDGs.

Table 11. Contribution of the LENOHA System to Sustainable Development Goals (SDGs)

| SDG | Goal Title | Contribution of LENOHA System | Specific Initiatives |
|---|---|---|---|
| SDG 3 | Good Health and Well-being | Improves equitable access and patient safety while reducing clinician burden. | Local, privacy-preserving Q&A for common pre-op queries; zero-hallucination FAQ answers; routing high-risk queries to clinicians. |
| SDG 4 | Quality Education | Enhances health literacy with consistent, comprehensible information. | Standardized patient-facing explanations; multilingual phrasing; staff on-boarding materials for consistent consent communication. |
| SDG 7 | Affordable and Clean Energy | Reduces energy use versus frontier LLMs and avoids energy-intensive retraining. | On-device inference; lightweight encoders; quantization; reporting Wh/request; scheduling batch jobs at low-carbon times (future work). |
| SDG 8 | Decent Work and Economic Growth | Improves efficiency and working conditions by offloading repetitive counseling. | Automated responses to routine FAQs; clinic-specific templates; dashboarding of outstanding non-routine queries. |
| SDG 9 | Industry, Innovation and Infrastructure | Promotes robust, locally deployable health AI infrastructure. | Open-source components; single-GPU deployment; offline mode for bandwidth-limited sites; IT-policy compliant logging/audit. |
| SDG 10 | Reduced Inequalities | Expands access in underserved regions and across abilities. | Local execution for remote islands; low-bandwidth UX; planned voice/assistive modes to support low vision users. |
| SDG 11 | Sustainable Cities and Communities | Supports resilient community healthcare and reduces unnecessary travel. | Pre-visit triage and education to cut avoidable referrals/returns; remote follow-up FAQs; printable care instructions. |
| SDG 12 | Responsible Consumption and Production | Encourages resource-efficient AI development and reuse. | Reuse of pre-trained models; modular FAQ curation instead of re-training; lifecycle governance for datasets and prompts. |
| SDG 13 | Climate Action | Lowers operational carbon footprint of patient-facing AI. | Energy-aware inference; on-premises power monitoring; target setting for Wh/request and periodic efficiency reviews. |

The LENOHA system not only aligns with the five core infrastructural requirements for ethical AI implementation proposed by Morley et al.[29] but also serves as a direct and timely response to the call for environmental sustainability and social justice recently articulated by Osmanlliu et al.[34]. Table 12 summarizes the alignment of the LENOHA system with the five core infrastructural requirements proposed by Morley et al.[29]. Therefore, our study presents a practical engineering blueprint that addresses these critical ethical and societal challenges and moves beyond narrow performance metrics.

**Table 12. Mapping the LENOHA System Contributions to the Ethical AI Implementation Requirements**

This table presents the alignment between the LENOHA system and the five foundational requirements for ethical AI implementation in healthcare proposed by Morley et al. Each requirement is matched with the corresponding design features of LENOHA, demonstrating how the system ensures robust data governance, epistemic integrity, value protection, accountability, and environmental sustainability. By satisfying all five pillars through deliberate architectural and operational choices, LENOHA exemplifies a responsible and ethically grounded model for AI deployment in clinical practice settings.

| Requirement No. | Ethical AI Implementation Requirement | LENOHA System's Corresponding Contribution |
|---|---|---|
| 1 | Robust Data Exchange | Ensure secure and interoperable data flow without reliance on external infrastructure. |
| 2 | Epistemic Certainty with Staff Autonomy | Establish a framework where medical staff remain the final decision-makers, acknowledging the limitations of AI. |
| 3 | Actively Protected Healthcare Values | Safeguard core healthcare values such as patient dignity, equity, and trust. |
| 4 | Validated Outcomes with Meaningful Accountability | Ensure transparency, traceability, and accountability for AI outputs. |
| 5 | Environmental Sustainability | Reduce environmental impact through energy-efficient AI deployment. |

Our research base is located in Nagasaki Prefecture, which is home to one of Japan's highest concentrations of remote islands. The challenges faced in accessing specialized medical services are not unique to Japan but are shared by remote communities worldwide, highlighting the global need for innovative solutions. As Udegbe et al. highlighted, the "digital divide "manifested in limited broadband access and low digital literacy, presents a major barrier to equitable healthcare in underserved regions [43].

This disparity is becoming increasingly pronounced with the advent of advanced medical AI. For instance, a recent nationwide study in the United States revealed that the adoption of medical AI is heavily concentrated in high-income metropolitan areas with academic medical centers, creating a risk of exacerbating existing healthcare inequalities[46]. The environmental footprint of computing is a cost driver with system-level implications for sustainability and resource allocation in healthcare. Notably, several evaluations label AI interventions as "dominant" or cost-saving while omitting indirect costs and upfront capital (e.g., infrastructure and integration) from the analysis. This mismatch risks overstating the net economic benefit of such system[3]

One key insight from this study is that the performance of clinical AI is not solely determined by computational scale. Instead, the irreplaceable value of human clinical knowledge—expert experiential knowledge rooted in practice and often inaccessible in web-derived corpora—remains fundamental to safety and reliability. In line with recent studies on clinical prediction, comparable task performance is achievable without frontier LLMs by appropriately structuring and leveraging this expertise[21],[8] [9], underscoring a knowledge-centric, complementary alternative to scale-driven development [48][7][27][21][9]

## 6 LIMITATIONS AND FUTURE WORK

This study had several limitations.

First, the system currently uses a text-based interface and does not support voice input, which constrains its applicability to some real-world dialogues. Simultaneously, in clinical settings, a text modality can confer practical advantages: (i) privacy—patients need not speak sensitive information in shared spaces; (ii) fidelity—technical terms and homophones can be conveyed precisely, avoiding safety risks from automatic speech recognition (ASR) errors; and (iii) psychological comfort—patients can compose and edit questions at their own pace, potentially reducing anxiety. These trade-offs motivate a focused question for future work: What is the optimal text/voice (or hybrid) interface for clinical environments?

Second, AI systems may inherit and amplify biases present in the training data. Our architecture mitigates this risk by using a high-precision sentence-transformer classifier for routing and by positioning AI strictly as a clinician support tool, rather than as a substitute. Nevertheless, residual bias remains a potential failure mode, warranting continuous monitoring and governance.

Third, our energy estimates integrated only the GPU board power; the CPU and system draw and facility-level overhead were not measured. Consistent with the notes in Table 10, our local figures reflect device-integral measurements; comprehensive accounting (CPU/system draw and facility overhead) was not included.

Fourth, both validation and test utterances were drafted by LLMs and then expert-reviewed and labeled; no real patient chat logs were used in this study. Therefore, the distribution shift to real-world phrasing remains a limitation that should be addressed prospectively.

Fifth, operating thresholds were fixed from validation via Youden's J; probability calibration and threshold-sensitivity analyses were not evaluated and remain to be performed.

Finally, although many reviews call for equitable and locally adapted AI, concrete implementation studies are rare. This study offers a practical response to this gap; however, further prospective evaluation is needed. As the next step,in a real-world, low-resource setting with the goal of advancing healthcare equity [30][23][39]. The system is also designed to improve accessibility for individuals with visual impairments, and we plan to conduct a systematic evaluation in future work.

## 7 DATA AVAILABILITY

The datasets generated and analyzed in this study, including the clinician-curated FAQ corpora tailored to the Japanese medical context, contain sensitive clinical content and cannot be publicly released under institutional and ethics committee regulations. We provide de-identified templates, schema definitions, prompts, and synthetic exemplar utterances that are sufficient to reproduce the pipeline. In addition, a limited subset of anonymized samples may be shared with qualified researchers for non-commercial research upon reasonable request to the corresponding author, subject to institutional review and a data-use agreement.

## 8 CODE AVAILABILITY

The code used in this study is available at: https://github.com/motokinaru/LENOHA-medical-dialogue. The repository includes configuration files, evaluation scripts, and prompts required to reproduce the results. The exact version used for this manuscript is tagged as a release in the repository, and an archival snapshot with a DOI will be deposited upon publication. Third-party model weights and datasets are subject to their respective license agreements.

White, Anders Andreassen, Tamara von Glehn, Lakshman Yagati, Mehran Kazemi, Lucas Gonzalez, Misha Khalman, Jakub Sygnowski, Alexandre Frechette, Charlotte Smith, Laura Culp, Lev Proleev, Yi Luan, Xi Chen, James Lottes, Nathan Schucher, Federico Lebron, Alban Rrustemi, Natalie Clay, Phil Crone, Tomas Kocisky, Jeffrey Zhao, Bartek Perz, Dian Yu, Heidi Howard, Adam Bloniarz, Jack W. Rae, Han Lu, Laurent Sifre, Marcello Maggioni, Fred Alcober, Dan Garrette, Megan Barnes, Shantanu Thakoor, Jacob Austin, Gabriel Barth-Maron, William Wong, Rishabh Joshi, Rahma Chaabouni, Deeni Fatiha, Arun Ahuja, Gaurav Singh Tomar, Evan Senter, Martin Chadwick, Ilya Kornakov, Nithya Attaluri, Iñaki Iturrate, Ruibo Liu, Yunxuan Li, Sarah Cogan, Jeremy Chen, Chao Jia, Chenjie Gu, Qiao Zhang, Jordan Grimstad, Ale Jakse Hartman, Xavier Garcia, Thanumalayan Sankaranarayana Pillai, Jacob Devlin, Michael Laskin, Diego de Las Casas, Dasha Valter, Connie Tao, Lorenzo Blanco, Adrià Puigdomènech Badia, David Reitter, Mianna Chen, Jenny Brennan, Clara Rivera, Sergey Brin, Shariq Iqbal, Gabriela Surita, Jane Labanowski, Abhi Rao, Stephanie Winkler, Emilio Parisotto, Yiming Gu, Kate Olszewska, Ravi Addanki, Antoine Miech, Annie Louis, Denis Teplyashin, Geoff Brown, Elliot Catt, Jan Balaguer, Jackie Xiang, Pidong Wang, Zoe Ashwood, Anton Briukhov, Albert Webson, Sanjay Ganapathy, Smit Sanghavi, Ajay Kannan, Ming-Wei Chang, Axel Stjerngren, Josip Djolonga, Yuting Sun, Ankur Bapna, Matthew Aitchison, Pedram Pejman, Henryk Michalewski, Tianhe Yu, Cindy Wang, Juliette Love, Junwhan Ahn, Dawn Bloxwich, Kehang Han, Peter Humphreys, Thibault Sellam, James Bradbury, Varun Godbole, Sina Samangooei, Bogdan Damoc, Alex Kaskasoli, Sébastien M. R. Arnold, Vijay Vasudevan, Shubham Agrawal, Jason Riesa, Dmitry Lepikhin, Richard Tanburn, Srivatsan Srinivasan, Hyeontaek Lim, Sarah Hodkinson, Pranav Shyam, Johan Ferret, Steven Hand, Ankush Garg, Tom Le Paine, Jian Li, Yujia Li, Minh Giang, Alexander Neitz, Zaheer Abbas, Sarah York, Machel Reid, Elizabeth Cole, Aakanksha Chowdhery, Dipanjan Das, Dominika Rogozińska, Vitaliy Nikolaev, Pablo Sprechmann, Zachary Nado, Lukas Zilka, Flavien Prost, Luheng He, Marianne Monteiro, Gaurav Mishra, Chris Welty, Josh Newlan, Dawei Jia, Miltiadis Allamanis, Clara Huiyi Hu, Raoul de Liedekerke, Justin Gilmer, Carl Saroufim, Shruti Rijhwani, Shaobo Hou, Disha Shrivastava, Anirudh Baddepudi, Alex Goldin, Adnan Ozturel, Albin Cassirer, Yunhan Xu, Daniel Sohn, Devendra Sachan, Reinald Kim Amplayo, Craig Swanson, Dessie Petrova, Shashi Narayan, Arthur Guez, Siddhartha Brahma, Jessica Landon, Miteyan Patel, Ruizhe Zhao, Kevin Villela, Luyu Wang, Wenhao Jia, Matthew Rahtz, Mai Giménez, Legg Yeung, James Keeling, Petko Georgiev, Diana Mincu, Boxi Wu, Salem Haykal, Rachel Saputro, Kiran Vodrahalli, James Qin, Zeynep Cankara, Abhanshu Sharma, Nick Fernando, Will Hawkins, Behnam Neyshabur, Solomon Kim, Adrian Hutter, Priyanka Agrawal, Alex Castro-Ros, George van den Driessche, Tao Wang, Fan Yang, Shuo-yiin Chang, Paul Komarek, Ross McIlroy, Mario Lučić, Guodong Zhang, Wael Farhan, Michael Sharman, Paul Natsev, Paul Michel, Yamini Bansal, Siyuan Qiao, Kris Cao, Siamak Shakeri, Christina Butterfield, Justin Chung, Paul Kishan Rubenstein, Shivani Agrawal, Arthur Mensch, Kedar Soparkar, Karel Lenc, Timothy Chung, Aedan Pope, Loren Maggiore, Jackie Kay, Priya Jhakra, Shibo Wang, Joshua Maynez, Mary Phuong, Taylor Tobin, Andrea Tacchetti, Maja Trebacz, Kevin Robinson, Yash Katariya, Sebastian Riedel, Paige Bailey, Kefan Xiao, Nimesh Ghelani, Lora Aroyo, Ambrose Slone, Neil Houlsby, Xuehan Xiong, Zhen Yang, Elena Gribovskaya, Jonas Adler, Mateo Wirth, Lisa Lee, Music Li, Thais Kagohara, Jay Pavagadhi, Sophie Bridgers, Anna Bortsova, Sanjay Ghemawat, Zafarali Ahmed, Tianqi Liu, Richard Powell, Vijay Bolina, Mariko Iinuma, Polina Zablotskaia, James Besley, Da-Woon Chung, Timothy Dozat, Ramona Comanescu, Xiance Si, Jeremy Greer, Guolong Su, Martin Polacek, Raphaël Lopez Kaufman, Simon Tokumine, Hexiang Hu, Elena Buchatskaya, Yingjie Miao, Mohamed Elhawaty, Aditya Siddhant, Nenad Tomasev, Jinwei Xing, Christina Greer, Helen Miller, Shereen Ashraf, Aurko Roy, Zizhao Zhang, Ada Ma, Angelos Filos, Milos Besta, Rory Blevins, Ted Klimenko, Chih-Kuan Yeh, Soravit Changpinyo, Jiaqi Mu, Oscar Chang, Mantas Pajarskas, Carrie Muir, Vered Cohen, Charline Le Lan, Krishna Haridasan, Amit Marathe, Steven Hansen, Sholto Douglas, Rajkumar Samuel, Mingqiu Wang, Sophia Austin, Chang Lan, Jiepu Jiang, Justin Chiu, Jaime Alonso Lorenzo, Lars Lowe Sjösund, Sébastien Cevey, Zach Gleicher, Thi Avrahami, Anudhyan Boral, Hansa Srinivasan, Vittorio Selo, Rhys May, Konstantinos Aisopos, Léonard Hussenot, Livio Baldini Soares, Kate Baumli, Michael B. Chang, Adrià Recasens, Ben Caine, Alexander Pritzel, Filip Pavetic, Fabio Pardo, Anita Gergely, Justin Frye, Vinay Ramasesh, Dan Horgan, Kartikeya Badola, Nora Kassner, Subhrajit Roy, Ethan Dyer,

Víctor Campos Campos, Alex Tomala, Yunhao Tang, Dalia El Badawy, Elspeth White, Basil Mustafa, Oran Lang, Abhishek Jindal, Sharad Vikram, Zhitao Gong, Sergi Caelles, Ross Hemsley, Gregory Thornton, Fangxiaoyu Feng, Wojciech Stokowiec, Ce Zheng, Phoebe Thacker, Çağlar Ünlü, Zhishuai Zhang, Mohammad Saleh, James Svensson, Max Bileschi, Piyush Patil, Ankesh Anand, Roman Ring, Katerina Tsihlas, Arpi Vezer, Marco Selvi, Toby Shevlane, Mikel Rodriguez, Tom Kwiatkowski, Samira Daruki, Keran Rong, Allan Dafoe, Nicholas FitzGerald, Keren Gu-Lemberg, Mina Khan, Lisa Anne Hendricks, Marie Pellat, Vladimir Feinberg, James Cobon-Kerr, Tara Sainath, Maribeth Rauh, Sayed Hadi Hashemi, Richard Ives, Yana Hasson, Eric Noland, Yuan Cao, Nathan Byrd, Le Hou, Qingze Wang, Thibault Sottiaux, Michela Paganini, Jean-Baptiste Lespiau, Alexandre Moufarek, Samer Hassan, Kaushik Shivakumar, Joost van Amersfoort, Amol Mandhane, Pratik Joshi, Anirudh Goyal, Matthew Tung, Andrew Brock, Hannah Sheahan, Vedant Misra, Cheng Li, Nemanja Rakićević, Mostafa Dehghani, Fangyu Liu, Sid Mittal, Junhyuk Oh, Seb Noury, Eren Sezener, Fantine Huot, Matthew Lamm, Nicola De Cao, Charlie Chen, Sidharth Mudgal, Romina Stella, Kevin Brooks, Gautam Vasudevan, Chenxi Liu, Mainak Chain, Nivedita Melinkeri, Aaron Cohen, Venus Wang, Kristie Seymore, Sergey Zubkov, Rahul Goel, Summer Yue, Sai Krishnakumaran, Brian Albert, Nate Hurley, Motoki Sano, Anhad Mohananey, Jonah Joughin, Egor Filonov, Tomasz Kępa, Yomna Eldawy, Jiawern Lim, Rahul Rishi, Shirin Badiezadegan, Taylor Bos, Jerry Chang, Sanil Jain, Sri Gayatri Sundara Padmanabhan, Subha Puttagunta, Kalpesh Krishna, Leslie Baker, Norbert Kalb, Vamsi Bedapudi, Adam Kurzrok, Shuntong Lei, Anthony Yu, Oren Litvin, Xiang Zhou, Zhichun Wu, Sam Sobell, Andrea Siciliano, Alan Papir, Robby Neale, Jonas Bragagnolo, Tej Toor, Tina Chen, Valentin Anklin, Feiran Wang, Richie Feng, Milad Gholami, Kevin Ling, Lijuan Liu, Jules Walter, Hamid Moghaddam, Arun Kishore, Jakub Adamek, Tyler Mercado, Jonathan Mallinson, Siddhinita Wandekar, Stephen Cagle, Eran Ofek, Guillermo Garrido, Clemens Lombriser, Maksim Mukha, Botu Sun, Hafeezul Rahman Mohammad, Josip Matak, Yadi Qian, Vikas Peswani, Pawel Janus, Quan Yuan, Leif Schelin, Oana David, Ankur Garg, Yifan He, Oleksii Duzhyi, Anton Älgmyr, Timothée Lottaz, Qi Li, Vikas Yadav, Luyao Xu, Alex Chinien, Rakesh Shivanna, Aleksandr Chuklin, Josie Li, Carrie Spadine, Travis Wolfe, Kareem Mohamed, Subhabrata Das, Zihang Dai, Kyle He, Daniel von Dincklage, Shyam Upadhyay, Akanksha Maurya, Luyan Chi, Sebastian Krause, Khalid Salama, Pam G Rabinovitch, Pavan Kumar Reddy M, Aarush Selvan, Mikhail Dektiarev, Golnaz Ghiasi, Erdem Guven, Himanshu Gupta, Boyi Liu, Deepak Sharma, Idan Heimlich Shtacher, Shachi Paul, Oscar Akerlund, François-Xavier Aubet, Terry Huang, Chen Zhu, Eric Zhu, Elico Teixeira, Matthew Fritze, Francesco Bertolini, Liana-Eleonora Marinescu, Martin Bölle, Dominik Paulus, Khyatti Gupta, Tejasi Latkar, Max Chang, Jason Sanders, Roopa Wilson, Xuewei Wu, Yi-Xuan Tan, Lam Nguyen Thiet, Tulsee Doshi, Sid Lall, Swaroop Mishra, Wanming Chen, Thang Luong, Seth Benjamin, Jasmine Lee, Ewa Andrejczuk, Dominik Rabiej, Vipul Ranjan, Krzysztof Styrc, Pengcheng Yin, Jon Simon, Malcolm Rose Harriott, Mudit Bansal, Alexei Robsky, Geoff Bacon, David Greene, Daniil Mirylenka, Chen Zhou, Obaid Sarvana, Abhimanyu Goyal, Samuel Andermatt, Patrick Siegler, Ben Horn, Assaf Israel, Francesco Pongetti, Chih-Wei "Louis" Chen, Marco Selvatici, Pedro Silva, Kathie Wang, Jackson Tolins, Kelvin Guu, Roey Yogev, Xiaochen Cai, Alessandro Agostini, Maulik Shah, Hung Nguyen, Noah Ó Donnaile, Sébastien Pereira, Linda Friso, Adam Stambler, Adam Kurzrok, Chenkai Kuang, Yan Romanikhin, Mark Geller, ZJ Yan, Kane Jang, Cheng-Chun Lee, Wojciech Fica, Eric Malmi, Qijun Tan, Dan Banica, Daniel Balle, Ryan Pham, Yanping Huang, Diana Avram, Hongzhi Shi, Jasjot Singh, Chris Hidey, Niharika Ahuja, Pranab Saxena, Dan Dooley, Srividya Pranavi Potharaju, Eileen O'Neill, Anand Gokulchandran, Ryan Foley, Kai Zhao, Mike Dusenberry, Yuan Liu, Pulkit Mehta, Ragha Kotikalapudi, Chalence Safranek-Shrader, Andrew Goodman, Joshua Kessinger, Eran Globen, Prateek Kolhar, Chris Gorgolewski, Ali Ibrahim, Yang Song, Ali Eichenbaum, Thomas Brovelli, Sahitya Potluri, Preethi Lahoti, Cip Baetu, Ali Ghorbani, Charles Chen, Andy Crawford, Shalini Pal, Mukund Sridhar, Petru Gurita, Asier Mujika, Igor Petrovski, Pierre-Louis Cedoz, Chenmei Li, Shiyuan Chen, Niccolò Dal Santo, Siddharth Goyal, Jitesh Punjabi, Karthik Kappaganthu, Chester Kwak, Pallavi LV, Sarmishta Velury, Himadri Choudhury, Jamie Hall, Premal Shah, Ricardo Figueira, Matt Thomas, Minjie Lu, Ting Zhou, Chintu Kumar, Thomas Jurdi, Sharat Chikkerur, Yenai Ma, Adams Yu, Soo Kwak, Victor Ähdel, Sujeevan Rajayogam, Travis Choma, Fei Liu, Aditya Barua, Colin Ji, Ji Ho Park, Vincent Hellendoorn, Alex Bailey, Taylan Bilal, Huanjie Zhou,

Mehrdad Khatir, Charles Sutton, Wojciech Rzadkowski, Fiona Macintosh, Roopali Vij, Konstantin Shagin, Paul Medina, Chen Liang, Jinjing Zhou, Pararth Shah, Yingying Bi, Attila Dankovics, Shipra Banga, Sabine Lehmann, Marissa Bredesen, Zifan Lin, John Eric Hoffmann, Jonathan Lai, Raynald Chung, Kai Yang, Nihal Balani, Arthur Bražinskas, Andrei Sozanschi, Matthew Hayes, Héctor Fernández Alcalde, Peter Makarov, Will Chen, Antonio Stella, Liselotte Snijders, Michael Mandl, Ante Kärrman, Paweł Nowak, Xinyi Wu, Alex Dyck, Krishnan Vaidyanathan, Raghavender R, Jessica Mallet, Mitch Rudominer, Eric Johnston, Sushil Mittal, Akhil Udathu, Janara Christensen, Vishal Verma, Zach Irving, Andreas Santucci, Gamaleldin Elsayed, Elnaz Davoodi, Marin Georgiev, Ian Tenney, Nan Hua, Geoffrey Cideron, Edouard Leurent, Mahmoud Alnahlawi, Ionut Georgescu, Nan Wei, Ivy Zheng, Dylan Scandinaro, Heinrich Jiang, Jasper Snoek, Mukund Sundararajan, Xuezhi Wang, Zack Ontiveros, Itay Karo, Jeremy Cole, Vinu Rajashekhar, Lara Tumeh, Eyal Ben-David, Rishub Jain, Jonathan Uesato, Romina Datta, Oskar Bunyan, Shimu Wu, John Zhang, Piotr Stanczyk, Ye Zhang, David Steiner, Subhajit Naskar, Michael Azzam, Matthew Johnson, Adam Paszke, Chung-Cheng Chiu, Jaume Sanchez Elias, Afroz Mohiuddin, Faizan Muhammad, Jin Miao, Andrew Lee, Nino Vieillard, Jane Park, Jiageng Zhang, Jeff Stanway, Drew Garmon, Abhijit Karmarkar, Zhe Dong, Jong Lee, Aviral Kumar, Luowei Zhou, Jonathan Evens, William Isaac, Geoffrey Irving, Edward Loper, Michael Fink, Isha Arkatkar, Nanxin Chen, Izhak Shafran, Ivan Petrychenko, Zhe Chen, Johnson Jia, Anselm Levskaya, Zhenkai Zhu, Peter Grabowski, Yu Mao, Alberto Magni, Kaisheng Yao, Javier Snaider, Norman Casagrande, Evan Palmer, Paul Suganthan, Alfonso Castaño, Irene Giannoumis, Wooyeol Kim, Mikołaj Rybiński, Ashwin Sreevatsa, Jennifer Prendki, David Soergel, Adrian Goedeckemeyer, Willi Gierke, Mohsen Jafari, Meenu Gaba, Jeremy Wiesner, Diana Gage Wright, Yawen Wei, Harsha Vashisht, Yana Kulizhskaya, Jay Hoover, Maigo Le, Lu Li, Chimezie Iwuanyanwu, Lu Liu, Kevin Ramirez, Andrey Khorlin, Albert Cui, Tian LIN, Marcus Wu, Ricardo Aguilar, Keith Pallo, Abhishek Chakladar, Ginger Perng, Elena Allica Abellan, Mingyang Zhang, Ishita Dasgupta, Nate Kushman, Ivo Penchev, Alena Repina, Xihui Wu, Tom van der Weide, Priya Ponnapalli, Caroline Kaplan, Jiri Simsa, Shuangfeng Li, Olivier Dousse, Fan Yang, Jeff Piper, Nathan Ie, Rama Pasumarthi, Nathan Lintz, Anitha Vijayakumar, Daniel Andor, Pedro Valenzuela, Minnie Lui, Cosmin Paduraru, Daiyi Peng, Katherine Lee, Shuyuan Zhang, Somer Greene, Duc Dung Nguyen, Paula Kurylowicz, Cassidy Hardin, Lucas Dixon, Lili Janzer, Kiam Choo, Ziqiang Feng, Biao Zhang, Achintya Singhal, Dayou Du, Dan McKinnon, Natasha Antropova, Tolga Bolukbasi, Orgad Keller, David Reid, Daniel Finchelstein, Maria Abi Raad, Remi Crocker, Peter Hawkins, Robert Dadashi, Colin Gaffney, Ken Franko, Anna Bulanova, Rémi Leblond, Shirley Chung, Harry Askham, Luis C. Cobo, Kelvin Xu, Felix Fischer, Jun Xu, Christina Sorokin, Chris Alberti, Chu-Cheng Lin, Colin Evans, Alek Dimitriev, Hannah Forbes, Dylan Banarse, Zora Tung, Mark Omernick, Colton Bishop, Rachel Sterneck, Rohan Jain, Jiawei Xia, Ehsan Amid, Francesco Piccinno, Xingyu Wang, Praseem Banzal, Daniel J. Mankowitz, Alex Polozov, Victoria Krakovna, Sasha Brown, MohammadHossein Bateni, Dennis Duan, Vlad Firoiu, Meghana Thotakuri, Tom Natan, Matthieu Geist, Ser tan Girgin, Hui Li, Jiayu Ye, Ofir Roval, Reiko Tojo, Michael Kwong, James Lee-Thorp, Christopher Yew, Danila Sinopalnikov, Sabela Ramos, John Mellor, Abhishek Sharma, Kathy Wu, David Miller, Nicolas Sonnerat, Denis Vnukov, Rory Greig, Jennifer Beattie, Emily Caveness, Libin Bai, Julian Eisenschlos, Alex Korchemniy, Tomy Tsai, Mimi Jasarevic, Weize Kong, Phuong Dao, Zeyu Zheng, Frederick Liu, Fan Yang, Rui Zhu, Tian Huey Teh, Jason Sanmiya, Evgeny Gladchenko, Nejc Trdin, Daniel Toyama, Evan Rosen, Sasan Tavakkol, Linting Xue, Chen Elkind, Oliver Woodman, John Carpenter, George Papamakarios, Rupert Kemp, Sushant Kafle, Tanya Grunina, Rishika Sinha, Alice Talbert, Diane Wu, Denese Owusu-Afriyie, Cosmo Du, Chloe Thornton, Jordi Pont-Tuset, Pradyumna Narayana, Jing Li, Saaber Fatehi, John Wieting, Omar Ajmeri, Benigno Uria, Yeongil Ko, Laura Knight, Amélie Héliou, Ning Niu, Shane Gu, Chenxi Pang, Yeqing Li, Nir Levine, Ariel Stolovich, Rebeca Santamaria-Fernandez, Sonam Goenka, Wenny Yustalim, Robin Strudel, Ali Elqursh, Charlie Deck, Hyo Lee, Zonglin Li, Kyle Levin, Raphael Hoffmann, Dan Holtmann-Rice, Olivier Bachem, Sho Arora, Christy Koh, Soheil Hassas Yeganeh, Siim Põder, Mukarram Tariq, Yanhua Sun, Lucian Ionita, Mojtaba Seyedhosseini, Pouya Tafti, Zhiyu Liu, Anmol Gulati, Jasmine Liu, Xinyu Ye, Bart Chrzaszcz, Lily Wang, Nikhil Sethi, Tianrun Li, Ben Brown, Shreya Singh, Wei Fan, Aaron Parisi, Joe Stanton, Vinod Koverkathu, Christopher A. Choquette-Choo, Yunjie Li,